\documentclass[10pt,twocolumn,letterpaper]{article}

\usepackage{cvpr}
\usepackage{times}
\usepackage{epsfig}
\usepackage{graphicx}
\usepackage{amsmath}
\usepackage{amssymb}
\usepackage{subfigure}

\usepackage[pagebackref=true,breaklinks=true,letterpaper=true,colorlinks,bookmarks=false]{hyperref}

\cvprfinalcopy 

\def\cvprPaperID{9448} 

\ifcvprfinal\fi
\begin{document}

\title{PCAMs: Weakly Supervised Semantic Segmentation Using Point Supervision}

\author{R. Austin McEver\hspace{0.1cm} \qquad\qquad B. S. Manjunath \\University of California, Santa Barbara}

\maketitle

\begin{abstract}
   Current state of the art methods for generating semantic segmentation rely heavily on a large set of images that have each pixel labeled with a class of interest label or background. Coming up with such labels, especially in domains that require an expert to do annotations, comes at a heavy cost in time and money. Several methods have shown that we can learn semantic segmentation from less expensive image-level labels, but the effectiveness of point level labels, a healthy compromise between all pixels labelled and none, still remains largely unexplored. This paper presents a novel procedure for producing semantic segmentation from images given some point level annotations. This method includes point annotations in the training of a convolutional neural network (CNN) for producing improved localization and class activation maps. Then, we use another CNN for predicting semantic affinities in order to propagate rough class labels and create pseudo semantic segmentation labels. Finally, we propose training a CNN that is normally fully supervised using our pseudo labels in place of ground truth labels, which further improves performance and simplifies the inference process by requiring just one CNN during inference rather than two. Our method achieves state of the art results for point supervised semantic segmentation on the PASCAL VOC 2012 dataset \cite{everingham2010pascal}, even outperforming state of the art methods for stronger bounding box and squiggle supervision.
\end{abstract}


\def \supfih {5.4cm}
\begin{figure}
    \centering
    \includegraphics[height=\supfih{}, width=0.45\linewidth]{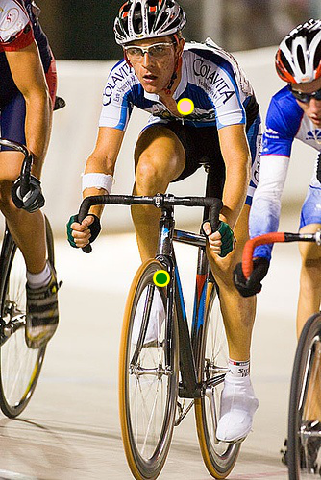}
    \includegraphics[height=\supfih{},width=0.45\linewidth]{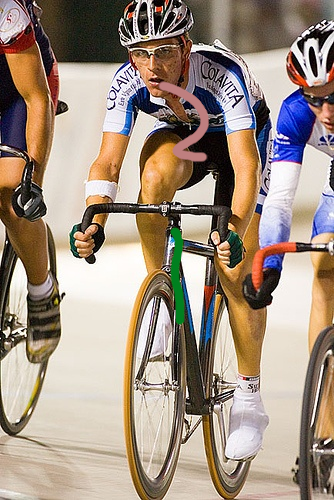}
    \includegraphics[height=\supfih{},width=0.45\linewidth]{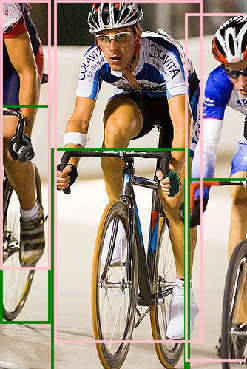}
    \includegraphics[height=\supfih{},width=0.45\linewidth]{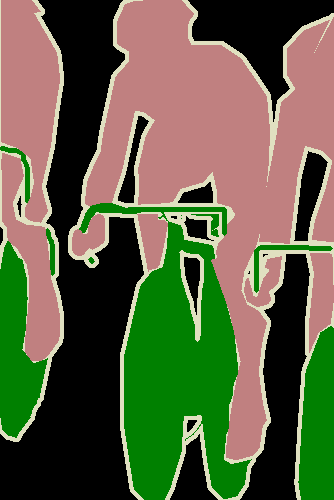}
    \caption{Top left: point annotations. Points are inflated and highlighted for visibility, but only the center pixel and its class label are collected and used. Top right: squiggle supervision. Bottom left: bounding box supervision. Bottom right: full supervision where white represents an ignore label that is not used during training and black is background. An image level label would only indicate that there are one or more instances of person and one or more instances of bicycle in this image.}
    \label{fig:intro}
\end{figure}

\section{Introduction}

Today's state of the art semantic segmentation algorithms generally rely on full supervision to achieve top performance. This means that annotators must label every pixel of every image in order to train their models. While this type of information is possible to obtain and available for common classes of interests (e.g. the person, dog, and cat class in natural image settings), many scientific applications cannot feasibly gather such detailed annotations because of the cost of annotator time, which may require a field expert. However, most settings can still afford some level of annotation.

\begin{figure*}
    \centering
    \includegraphics[width=1\linewidth]{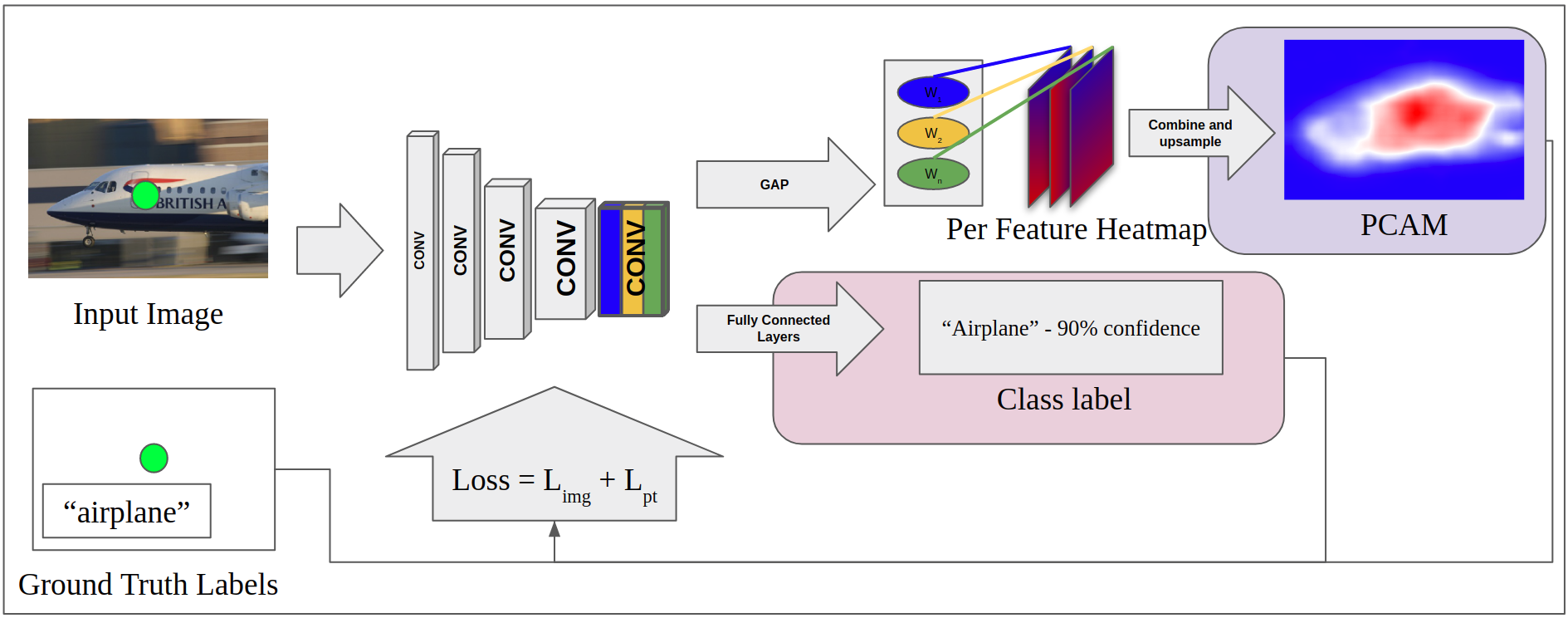}
    \caption{PCAM training overview: compute PCAM with a point supervised input image using the ResNet50 \cite{he2016deep} backbone, produce the class labels, and compare the output with the supervised point and class labels to compute a loss used to update the PCAM network. Note that the last convolutional layer's feature maps are colored to match with their class dependent weight vector to produce per feature heatmaps, which are then combined and upsampled during training to generate the PCAM.}
    \label{fig:method}
\end{figure*}

This problem has motivated the exploration of weakly supervised semantic segmentation, especially the setting where image level labels are used for training \cite{ahn2019weakly,ahn2018learning, briq2018convolutional, wang2018weakly, wei2018revisiting, ge2018multi, hong2017weakly, lee2019ficklenet, huang2018weakly, wei2017object, durand2017wildcat, roy2017combining, kwak2017weakly, lu2016learning, liang2016learning, kolesnikov2016seed, pathak2015constrained,pinheiro2015image}. We make the distinction for image level labels because there has also been some attention where bounding boxes \cite{song2019box, hu2018learning, zhao2018pseudo, khoreva2017simple, papandreou2015weakly, dai2015boxsup, rajchl2016deepcut, maninis2018deep}, points \cite{bearman2016s}, scribbles \cite{vernaza2017learning, tang2018normalized, lin2016scribblesup, xu2015learning} or other forms of supervision \cite{pathak2015constrained} have been used for the weakly supervised segmentation problem. Figure \ref{fig:intro} shows visual examples of a few of these common types of weak supervision for the semantic segmentation problem. 

Hence, we consider weak supervision to include any supervised learning algorithm that makes finer predictions than the annotations from which it should learn. For example, a method for solving image level supervised semantic segmentation (the most common form of weak supervision for segmentation), uses image level labels (i.e. simple labels of which classes appear in each training image) to train and then produces semantic segmentation, which is a labelling of each pixel as a particular class of interest or background. A large gap still remains in performance between algorithms using only image level labels and their fully supervised counterparts with more supervision generally providing better performance.

While focusing on image level supervision makes sense for many cases, such as those where many images of classes of interest can be scraped from the web \cite{hong2017weakly, liang2016learning}, not all problems are that way. Many research groups in more scientific fields heavily guard their images and annotations such that a given group might only have access to its own private dataset. Further, these datasets can be much more expensive to annotate than natural images, requiring field experts to recognize and localize classes of interest. Therefore, it is imperative that we develop methods that make optimal use of annotator time. Bearman, et al \cite{bearman2016s} showed that, in their setting, point supervision returned the best results given a fixed budget of annotator time. 

In this paper, we refer to point supervision as the same setting as used presented by Bearman et al \cite{bearman2016s}. Specifically, we are interested in the setting where annotators view an image and annotate one or more instances of each class of interest in the image. This setting makes sense when users are manually annotating images as clicking on each instance takes little time beyond providing image level labels and provides valuable localization information that can be used with this paper's method and ultimately leads to significantly better segmentation results. Further, with large or dense images, this setting might even help annotators keep track of which objects have already been noted and which have not. During training, then, models would not have access to any information beyond point annotations. 

Despite its effectiveness and efficiency, this type of point supervision for semantic segmentation has not been explored in several years since Bearman et al \cite{bearman2016s}.

In this paper, we present a novel method for achieving state of the art point-supervised semantic segmentation. At the center of our method, we utilize point localization information during the training of class activation maps (CAMs) \cite{zhou2016learning}. We then use IRNet  \cite{ahn2019weakly} to refine these point supervised class activation maps (PCAMs) to create pseudo semantic segmentation labels, which are then used as ground truth for training a fully supervised semantic segmentation network.

The primary contributions of this paper are as follows:
\begin{itemize}
  \item We introduce a novel method for including point supervision in the training of CAMs to produce PCAMs, discussed in Section 3.1.
  \item We achieve state of the art performance on the PASCAL VOC 2012 dataset \cite{everingham2010pascal} for point supervised semantic segmentation, outperforming older fully supervised methods and current state of the art methods utilizing even more supervision such as bounding boxes and squiggles.
\end{itemize}
Our code will be made available on Github.

\section{Related Work}

This section reviews several methods on which our work is based. First, we review some methods for coarse localization given weak labels. Then we discuss some state of the art image-level supervised segmentation papers before finally reviewing point-level supervised segmentation works.

\subsection{Weakly Supervised Object Localization}
Weakly supervised localization refers to the problem of finding areas of interest in an image given weak labels. This problem has also been referred to as saliency detection. These methods \cite{itti1998model, hou2007saliency, alexe2012measuring} attempt to find class generic regions that are likely to contain some class of interest using varying levels of supervision. Similarly, methods such as the one presented by Cholakka et al \cite{cholakkal2016backtracking} work to localize areas that are likely to contain a specific classes of interest. 

Following the path of several methods which use CNNs for weakly supervised object localization \cite{bazzani2016self, oquab2015object, cinbis2016weakly, oquab2014learning}, CAMs as created by Zhou et al \cite{zhou2016learning} have become the basis for many weakly supervised class specific localization problems including some of the methods that will be discussed in Section 2.2 \cite{ahn2019weakly, ahn2018learning, wei2017object}. 

In short, Zhou et al \cite{zhou2016learning} propose a method for interpreting the activation of a CNN trained for classification that provides localization information for the classes of interest. This localization is done by adding a global average pooling (GAP) layer to a classification network and examining the activations of the final convolutional layer in the classification CNN. The CNN has class dependent weight vectors that learn which activations correspond with each class. By weighting the feature map of the last convolutional layer in the CNN and upsampling the weighted map to the image size, the authors can begin to localize which areas of an image correspond to different classes. Section 3.1 further discusses this method.

Ultimately, this process allows for coarse localization of classes of interest that can be derived from a network trained with only image level labels. Other methods have built on CAMs for creating even better class specific localization extended beyond the image domain \cite{selvaraju2017grad, chattopadhay2018grad}. Zhang et al \cite{zhang2018adversarial} proposed a method for iteratively improving localization performance by erasing areas of high activation from training images, forcing the CNN to learn other features associated with classes of interest thereby expanding areas of localization.

Still, this line of work is only able to achieve coarse localization, and the resultant activation maps rarely cover the full extent of classes of interest. These activation maps also correspond poorly with object boundaries.

\subsection{Image-level Supervised Segmentation}

Given the ease of collecting image level labels, many efforts have recently been made to create effective segmentation algorithms that use only these labels \cite{ahn2019weakly,ahn2018learning, briq2018convolutional, wang2018weakly, wei2018revisiting, ge2018multi, hong2017weakly, lee2019ficklenet, huang2018weakly, wei2017object, durand2017wildcat, roy2017combining, kwak2017weakly, lu2016learning, liang2016learning, kolesnikov2016seed, pathak2015constrained,pinheiro2015image}. 

Kolesnikov et al \cite{kolesnikov2016seed} propose a method for training a segmentation CNN with a loss that guides the network via its loss function to follow localization cues, expand around those cues, and adhere to object boundaries. Several methods follow a similar path including work done by Huang et al \cite{huang2018weakly} which uses CAMs to generate its localization cues. Similarly, their loss function has a term for adhering to these cues, growing their regions according to some similarity criteria, and adhering to object boundaries. 

The most relevant methods to this paper include recent works of Ahn et al \cite{ahn2019weakly, ahn2018learning} which also generally rely on the method of Zhou et al \cite{zhou2016learning} to create CAMs. Ahn's method generates CAMs, mines affinity labels from the CAMs, and presents a neural network that outputs information that can be used to generate a transition probability matrix. In one of their works \cite{ahn2018learning}, Ahn et al describe a method for carefully exploiting CAMs to generate positive and negative affinity labels for pixel pairs. Positive affinity labels should indicate a pair of pixels is of the same class while negative labels indicates differing classes. These labels can be generated automatically by giving thresholds for confidence in CAMs such that two pixels that have high confidence in the same class are assigned a positive affinity label. 

Each work by Ahn et al presents different networks and methods for learning from these mined affinity labels, but they both output information that is ultimately used to generate a transition probability matrix. They then perform a random walk over CAMs using the computed transition probabilities to propagate class labels and refine CAMs into pseudo segmentation labels. Finally, these pseudo labels are used in place of ground truth to train a CNN that is designed to perform segmentation given real ground truth labels.

We largely follow this type of method with the novel introduction of point supervision during early stages that leads to a significant improvement in performance.

\subsection{Point-level Supervised Segmentation}

While image level supervision certainly has its place in domains where images can easily be collected for classes of interest, point supervision can significantly improve performance of segmentation algorithms and can easily be collected when annotating a new dataset. To our knowledge, no publications have tackled the problem of point supervised segmentation algorithms since Bearman et al \cite{bearman2016s} introduced the problem, collected and provided point level annotations for the PASCAL VOC 2012 dataset \cite{everingham2010pascal}, and proposed a method for incorporating point supervision and localization cues into training a normally fully supervised network directly. This method computes a loss over supervised points that is based on a log softmax probability. 

The work by Mainis et al \cite{maninis2018deep} is sometimes mentioned in literature as using point level annotations to achieve semantic segmentation; however, this paper uses extreme points rather than more random points as collected in Bearman's work where annotators are simply asked to click on objects of interest\cite{bearman2016s}. 

Rather than point-level annotations, then, \cite{maninis2018deep} more closely resembles bounding box level supervision. Arguably, this method uses even more supervision than bounding boxes in that extreme points give bounding box information in addition to four points that are certainly on the boundary of a given object. When given only a bounding box, it is uncertain which parts of the bounding box edges actually lie on an object boundary.

\section{Framework}
The primary contribution of this paper is to introduce a method for training PCAMs: point supervised class activation maps as shown in Figure \ref{fig:method}. Including point supervision in the training process in this way significantly improves the localization performance of CAMs. To achieve state of the art semantic segmentation results, we closely follow the method of \cite{ahn2019weakly} to train IRNet on our PCAMs and use its output to refine PCAMs to create pseudo semantic segmentation labels. Finally, we train the fully supervised segmentation network  presented in \cite{chen2018encoder}, DeepLabv3+, on the pseudo labels and use this network for final predictions.

\def \camfih {3.0cm}

\begin{figure*}[ht!]
 
\begin{subfigure}{}
\includegraphics[width=0.245\linewidth, height=\camfih{}]{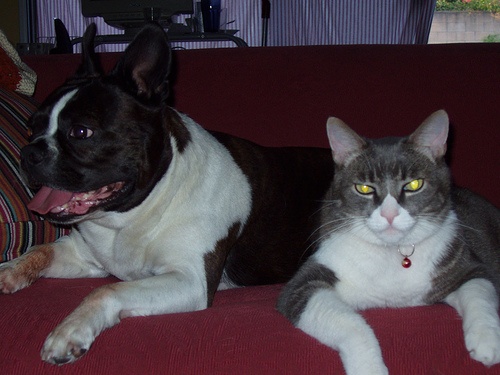} 
\includegraphics[width=0.245\linewidth, height=\camfih{}]{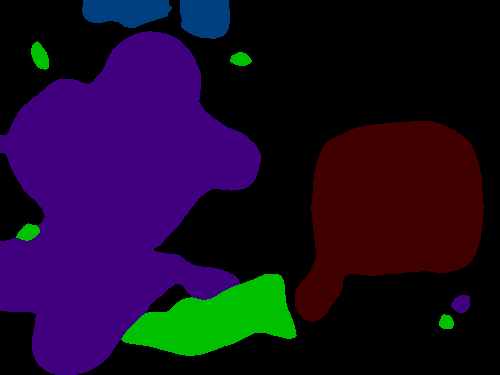} 
\includegraphics[width=0.245\linewidth, height=\camfih{}]{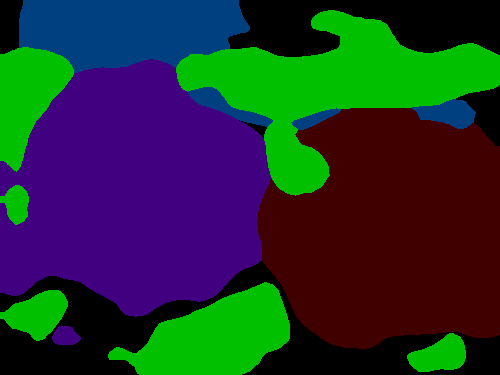} 
\includegraphics[width=0.245\linewidth, height=\camfih{}]{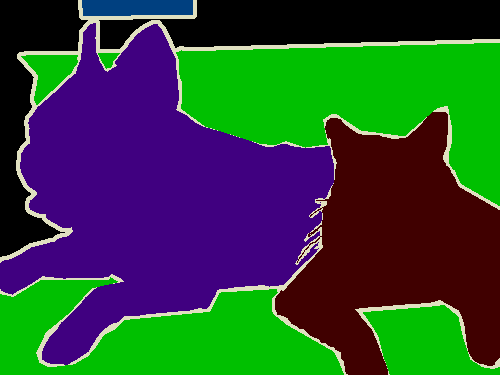}
\label{fig:subim1}
\end{subfigure}

\begin{subfigure}{}
\includegraphics[width=0.245\linewidth, height=\camfih{}]{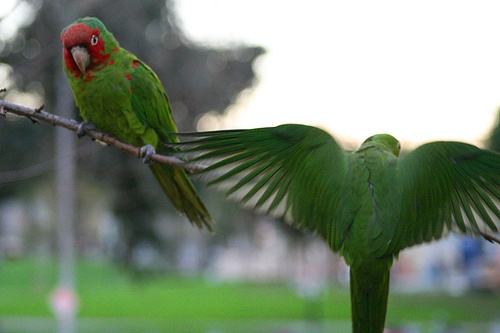} 
\includegraphics[width=0.245\linewidth, height=\camfih{}]{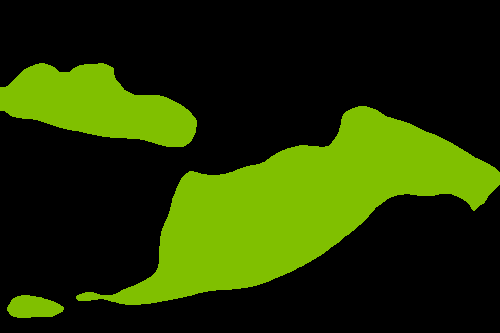} 
\includegraphics[width=0.245\linewidth, height=\camfih{}]{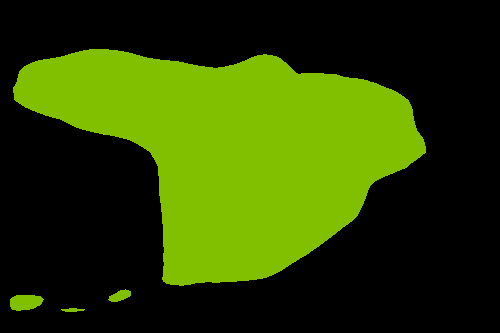} 
\includegraphics[width=0.245\linewidth, height=\camfih{}]{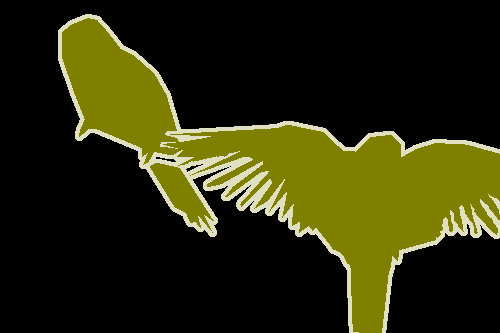}
\label{fig:subim2}
\end{subfigure}

\begin{subfigure}{}
\includegraphics[width=0.245\linewidth, height=\camfih{}]{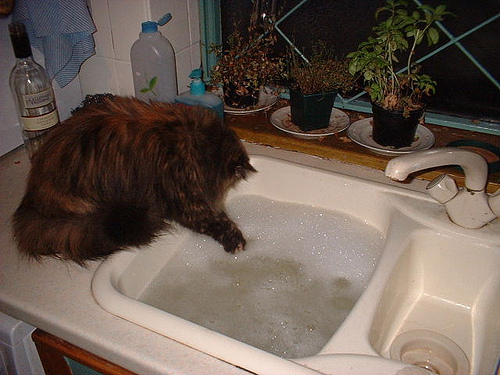} 
\includegraphics[width=0.245\linewidth, height=\camfih{}]{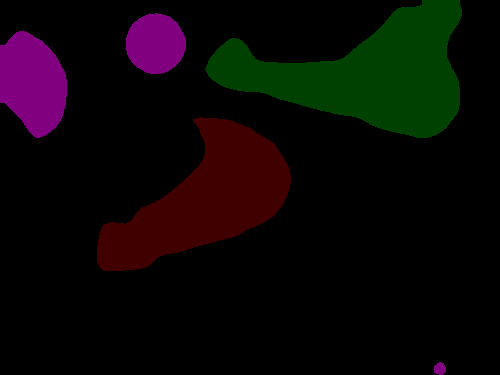} 
\includegraphics[width=0.245\linewidth, height=\camfih{}]{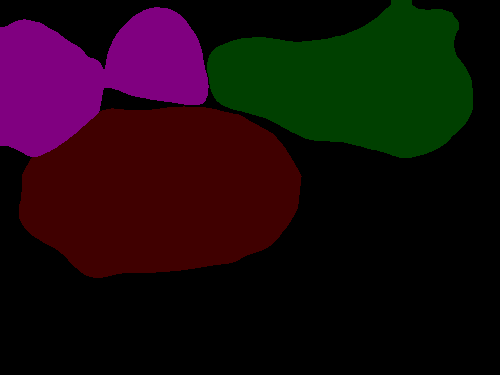} 
\includegraphics[width=0.245\linewidth, height=\camfih{}]{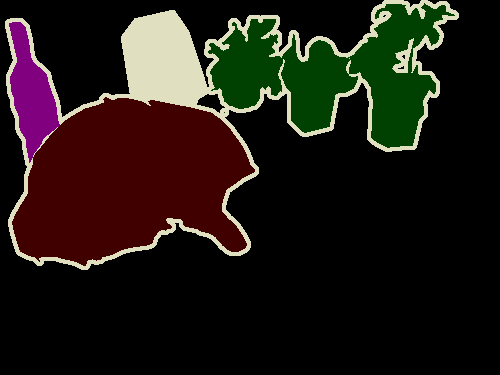}
\label{fig:subim3}
\end{subfigure}
 
\caption{From left to right: original images, CAM labels, PCAM labels, and ground truth segmentation for each image}
\label{fig:camgrid}
\end{figure*}

\subsection{PCAMs}

To train PCAMs we generally follow the method presented in the work by Zhou et al \cite{zhou2016learning} but present an additional step that allows the inclusion of point supervision during training as shown in Figure \ref{fig:method}. The original method for producing CAMs adds a global average pooling layer to an image classification CNN. Following the work by Ahn et al \cite{ahn2019weakly}, we use ResNet50 \cite{he2016deep} as the backbone classification network and drop the stride of its last downsampling layer to prevent too much reduction in resolution. Training the image classification CNN is done with a simple multilabel soft margin loss.

\begin{align} \label{eq:limg}
    L_{img}(\hat{y},y) = -\frac{1}{C}*\sum_{i=1}^{C}(y_{i}*\log(  (  1+\exp(-\hat{y}_{i})  )^{-1}  ) \\
    + (1-y_{i})*\log(\frac{\exp(-\hat{y}_{i})}{1+\exp(-\hat{y}_{i})})) \notag
\end{align}

where $y$ is the one hot encoded vector of classes in the images, $\hat{y}$ is the predicted class vector, $y_{i}$ is the label of the $i$th class, and $C$ is the number of classes. Once trained, CAMs of a given class c can be generated by

\begin{equation}
    M_{c}(\mathbf{x}) = \frac{\phi_{c}^{\top}f(\mathbf{x})}{ max_{\mathbf{x}} \phi_{c}^{\top}f(\mathbf{x}) }
\label{eq:cam}
\end{equation}

Where $\phi_{c}$ represents the learned class-dependent weight vector for class $c$, $f$ is the feature map from the final convolutional layer of the classification network backbone, and $x$ is a 2D coordinate in $f$. As mentioned before, we use ResNet50 \cite{he2016deep} as our backbone network with a reduced stride in the final downsampling layer. The dimensions of the CAMs are then 1/16 of the input image. 

Training the classification network with point annotations, however, requires that we generate the CAMs during training. To align the CAMs with the input image, we use bilinear interpolation to upsample the CAMs to the size of the input image to create $U_{c}$, the upsampled CAM for class $c$. This upsampling is only done during inference in previous works, but our method introduces this upsampling during training so that we can compare the upsampled CAM with any supervised points and guide the network for better activation mapping. We then compute the mean squared error loss over each of the supervised points as follows:

\begin{equation}
    L_{pt}^{c} = \frac{1}{|S|}\sum_{s \epsilon S} (U_{c}(s) - G_{c}(s))^{2}
\label{eq:lptc}
\end{equation}

where $S$ is the set of supervised pixel locations, $U_{c}(s)$ is the predicted probability that location $s$ is of class $c$, and $G_{c}(s)$ is the binary ground truth label for class $c$ for the pixel at location $s$. For the point supervised term of the loss for training a network to generate PCAMs, we average the classwise losses for each class present in the training image.

\begin{equation}
L_{pt} = \frac{1}{|C{}'|}*\sum_{c\epsilon C{}'}L_{pt}^{c}
\label{eq:lpt}
\end{equation}

where $C{}'$ is the set of classes in the training image. This allows us to precisely use any point level annotations to guide the network to activate at specific locations. This loss term also leads to more confident activation maps that cover a greater spatial extent of objects of interest. The total loss for training the PCAM network is then

\begin{equation}
L = L_{img} + \alpha L_{pt}
\label{eq:loss}
\end{equation}

where $\alpha$ is a weighting term.

\subsection{IRNet}

We follow the work of Ahn et al \cite{ahn2019weakly} in training and using IRNet. We mine semantic affinity labels from our improved PCAMs, which use point level supervision, rather than mining labels from CAMs which use image level supervision, and use them for training IRNet. In order to mine semantic affinity labels, we threshold each PCAM such that low values are considered background and high values are considered as the corresponding class. We ignore all pixels that have middling confidence values as their labels are uncertain and affinity labels must be mined reliably to optimize the performance of IRNet.

We then examine all pixels within a small radius of confident pixels. If a pair of pixels are both confident or background and have the same label, the pair is assigned a positive affinity label, and if it has a different label it is assigned a negative affinity label.

IRNet uses training images and these mined affinity labels to learn to predict a displacement vector field and class boundary map. The displacement field should indicate centroids of class boundaries, which aims to segment class instance but also aids in separating instances of adjacent differing classes. The class boundary map aims to indicate boundaries of classes. 

The class boundary map can be synthesized to create an affinity matrix. If a pair of pixels have a positive class boundary pixel on the line between them, they have low semantic affinity. Otherwise, they are likely to be of the same class and have high semantic affinity. Next, the semantic affinities are used to compute a transition probability matrix for a random walk which is performed over instance-wise PCAMs. Guided by this transition probability matrix, a random walk over PCAMs expands and refines activation areas. These refined PCAMs are then combined to create pseudo semantic segmentation labels.

\subsection{DeepLabv3+}

The final step in our method takes the pseudo semantic segmentation labels generated via random walk propagation over PCAMs and uses them as ground truth for training DeepLabv3+ \cite{chen2018encoder}. We find that, despite training on imperfect labels, the network is still able to generate slightly better segmentation results on unseen images when compared to refined PCAMs. 

Further, using a trained fully supervised network makes inference simpler. To infer a new image otherwise, we would need to first run inference using the trained PCAM network and run inference using IRNet before we could compute the transition probability network from IRNet's output. Finally, we would need to run the random walk algorithm on the PCAM using the generated transition probability matrix to refine the PCAM. 

After we train DeepLabv3+ on our pseudo semantic segmentation labels, inference on an unseen image can be done simply by running inference with our trained DeepLabv3+ model.

\section{Experiments and Results}
The following section describes our experiments on the PASCAL VOC 2012 dataset \cite{everingham2010pascal} on which we achieve state of the art performance for point-supervised semantic segmentation.

\subsection{Dataset}
All of our experimental results are reported on the PASCAL VOC 2012 dataset \cite{everingham2010pascal} and trained using the PASCAL VOC 2012 training images supplemented with the images from the SBD dataset \cite{hariharan2011semantic}, following common practice \cite{bearman2016s, ahn2019weakly, ahn2018learning}. The PASCAL VOC 2012 dataset includes 1,464 training images and 1,449 validatation images. The SBD dataset contains annotations for 11,355 images from the PASCAL VOC 2012 dataset. In total, we train with 10,582 training images and test with 1,449 validation images.

For training PCAMs, we use the point level labels provided by \cite{bearman2016s}. These labels include one ore more annotated points per class in each training image. Overall, we use an average of approximately 2.4 points per image for training.

\subsection{Hyperparameters}

\subsubsection{PCAM Network}
The PCAM network uses ResNet50 \cite{he2016deep} as its backbone network. The learning rate is initially set to 0.001 for the backbone parameters and 0.01 for the classification layers. The loss weighting term $\alpha$ is set to 0.1.

\subsubsection{IRNet}
We generally use IRNet as presented by Ahn et al \cite{ahn2019weakly}. We set the CAM evaluation threshold for producing the labels for IRNet to 0.3. Similarly, we set the threshold for semantic segmentation at 0.3. These adjustments compensate for PCAMs being somewhat more confident than CAMs. For our setting, we had slightly better results setting IRNet's $\beta$ parameter to 12. This parameter affects the generation of the transition matrix that is used for the random walk propogation of attention scores and is detailed in the paper by Ahn et al\cite{ahn2019weakly}.

\begin{table}
\begin{center}
  \begin{tabular}{l |c| c | c }
    \hline
    Activation Map      & Sup. &  \textit{train}  & \textit{val} \\ \hline
    CAM      & \textit{I} & 48.3 &  46.0 \\ \hline
    PCAM & \textit{P} & 56.5 & 54.4 \\ \hline
    PCAM & \textit{F} & 64.0 & 55.5 \\ \hline
  \end{tabular}
\end{center}
\caption{mIOU comparison of CAM, PCAM, and PCAM trained with all points on the PASCAL VOC 2012 \textit{train} and \textit{val} sets 
}
\label{tab:CAM}
\end{table}

\begin{table}
\begin{center}
  \begin{tabular}{l | c | c}
    \hline
    Refined Activation Map      &  \textit{train}  & \textit{val} \\ \hline
    CAM        & 66.5 &  57.4 \\ \hline
    PCAM & 70.8 & 68.5 \\ \hline
  \end{tabular}
\end{center}
\caption{mIOU comparison of performance on PASCAL VOC 2012 \textit{train} and \textit{val} sets of CAMs and PCAMs after being refined by random walk via IRNet's transition matrix
}
\label{tab:refined}
\end{table}

\begin{figure*}[]
 
\begin{subfigure}{}
\includegraphics[width=0.16\linewidth, height=3.5cm]{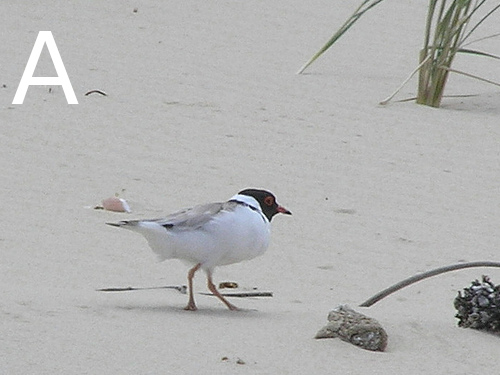} \includegraphics[width=0.16\linewidth, height=3.5cm]{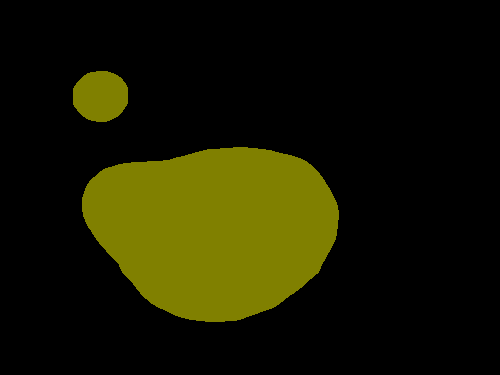} 
\includegraphics[width=0.16\linewidth, height=3.5cm]{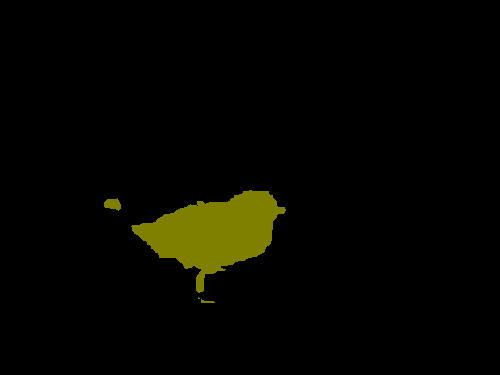} 
\includegraphics[width=0.16\linewidth, height=3.5cm]{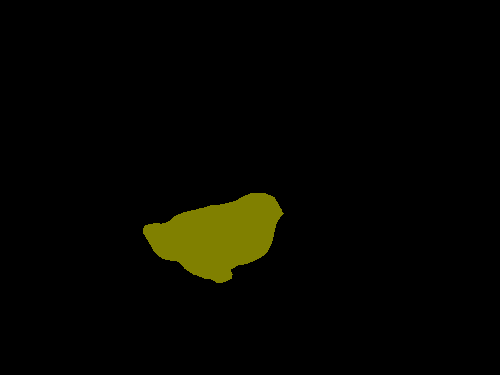}
\includegraphics[width=0.16\linewidth, height=3.5cm]{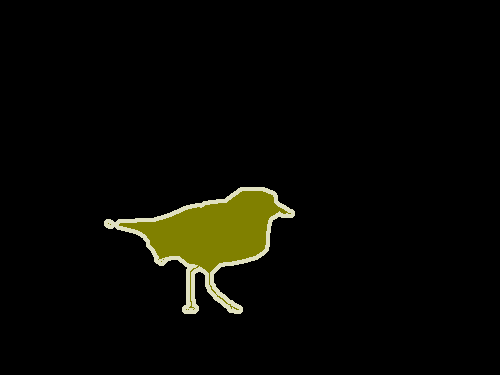}
\includegraphics[width=0.16\linewidth, height=3.5cm]{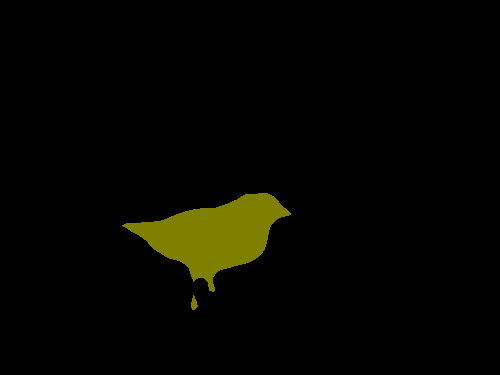}
\label{fig:subim10}
\end{subfigure}

\begin{subfigure}{}
\includegraphics[width=0.16\linewidth, height=3.5cm]{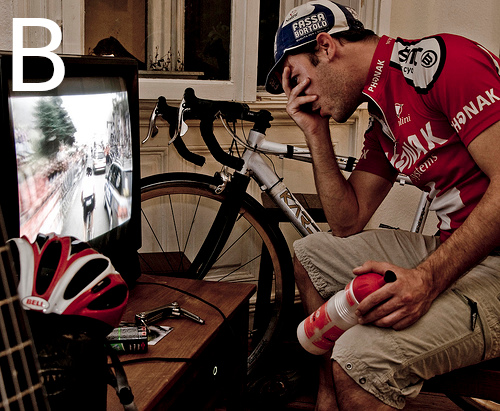} 
\includegraphics[width=0.16\linewidth, height=3.5cm]{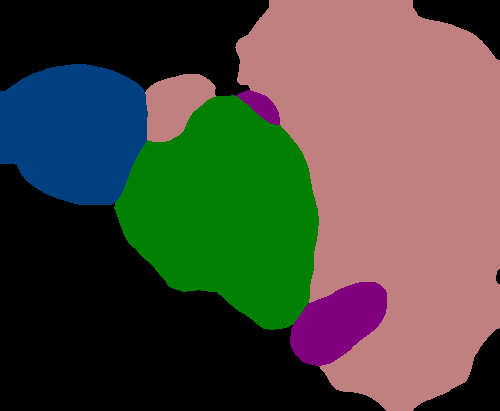} 
\includegraphics[width=0.16\linewidth, height=3.5cm]{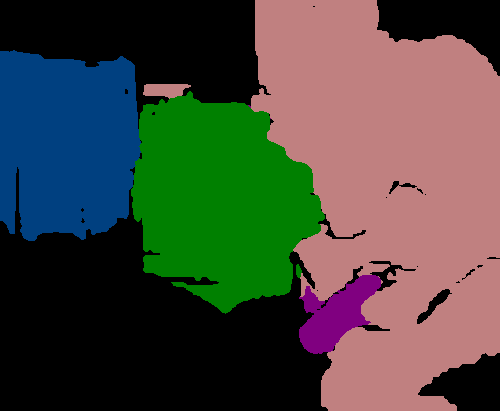} 
\includegraphics[width=0.16\linewidth, height=3.5cm]{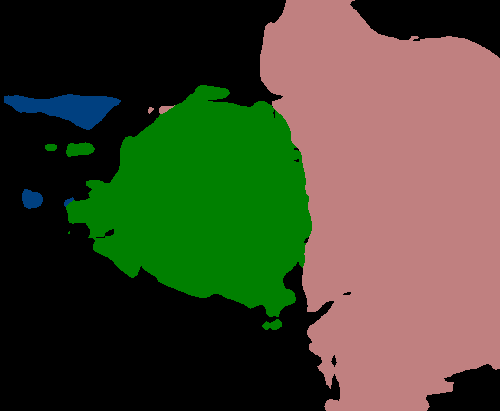}
\includegraphics[width=0.16\linewidth, height=3.5cm]{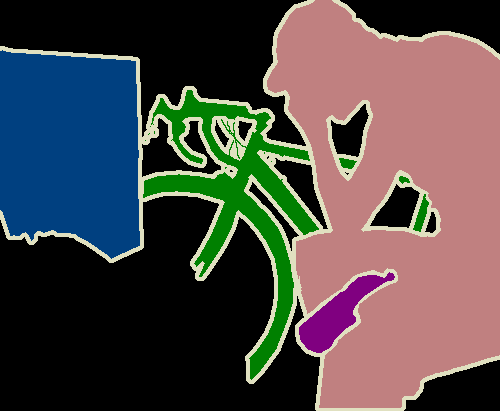}
\includegraphics[width=0.16\linewidth, height=3.5cm]{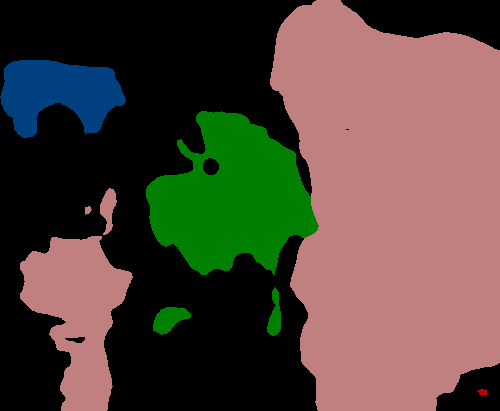}
\end{subfigure}

\begin{subfigure}{}
\includegraphics[width=0.16\linewidth, height=3.5cm]{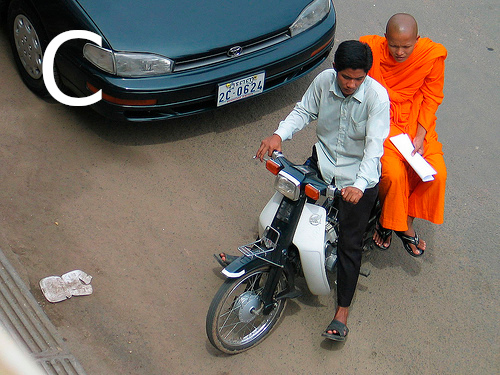} 
\includegraphics[width=0.16\linewidth, height=3.5cm]{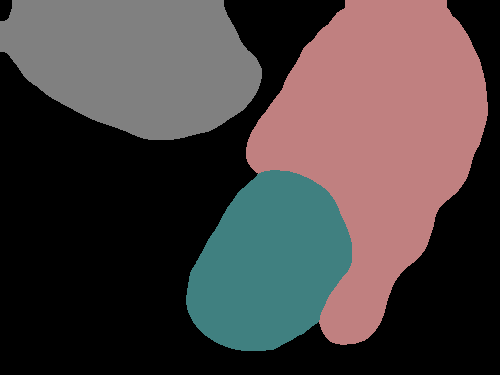} 
\includegraphics[width=0.16\linewidth, height=3.5cm]{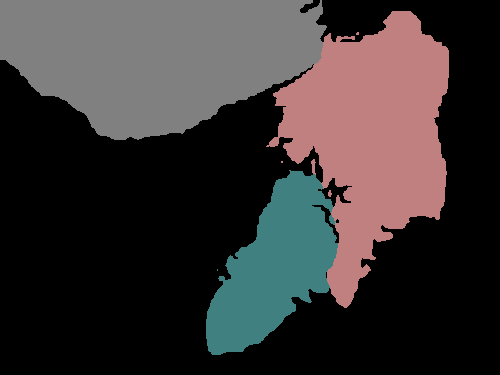} 
\includegraphics[width=0.16\linewidth, height=3.5cm]{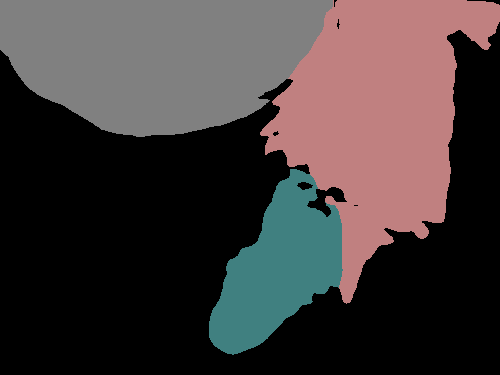}
\includegraphics[width=0.16\linewidth, height=3.5cm]{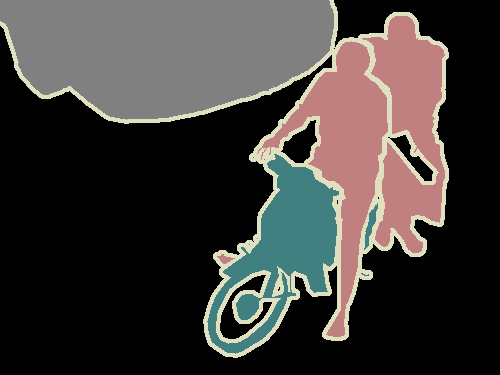}
\includegraphics[width=0.16\linewidth, height=3.5cm]{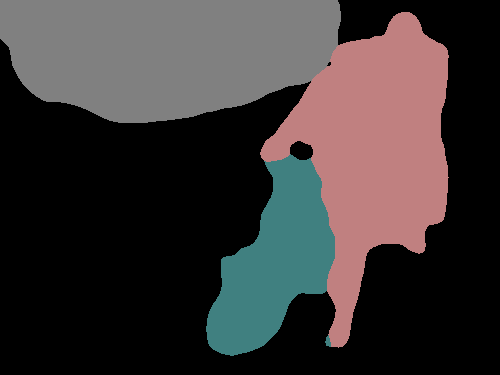}
\end{subfigure}

\begin{subfigure}{}
\includegraphics[width=0.16\linewidth, height=3.5cm]{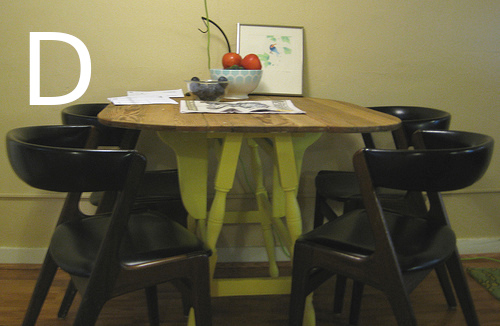} 
\includegraphics[width=0.16\linewidth, height=3.5cm]{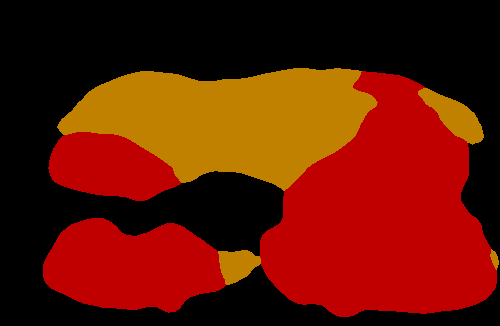} 
\includegraphics[width=0.16\linewidth, height=3.5cm]{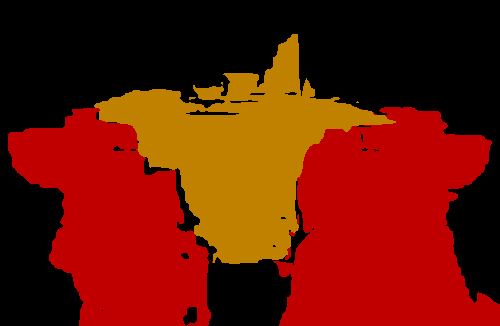} 
\includegraphics[width=0.16\linewidth, height=3.5cm]{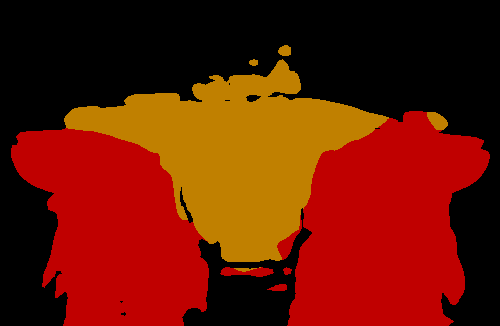}
\includegraphics[width=0.16\linewidth, height=3.5cm]{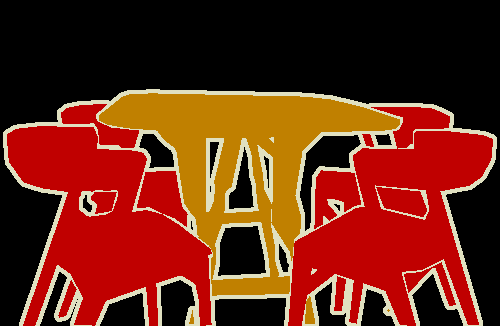}
\includegraphics[width=0.16\linewidth, height=3.5cm]{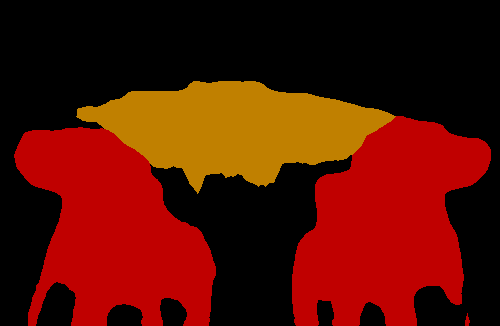}
\end{subfigure}

\begin{subfigure}{}
\includegraphics[width=0.16\linewidth, height=3.5cm]{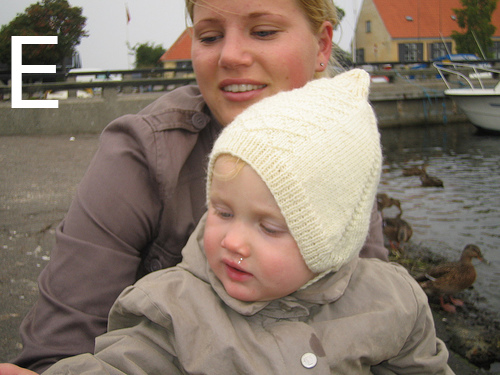} 
\includegraphics[width=0.16\linewidth, height=3.5cm]{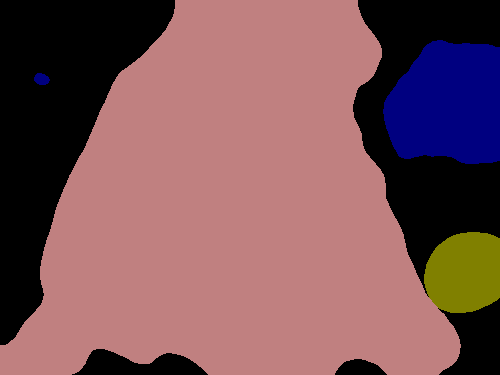} 
\includegraphics[width=0.16\linewidth, height=3.5cm]{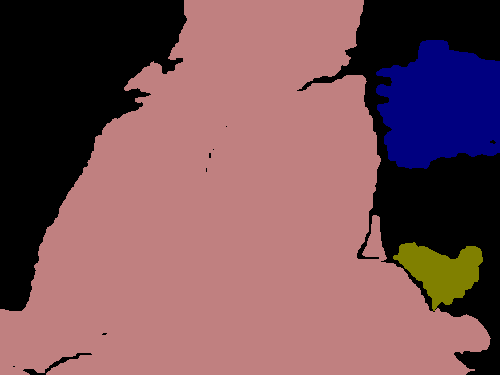} 
\includegraphics[width=0.16\linewidth, height=3.5cm]{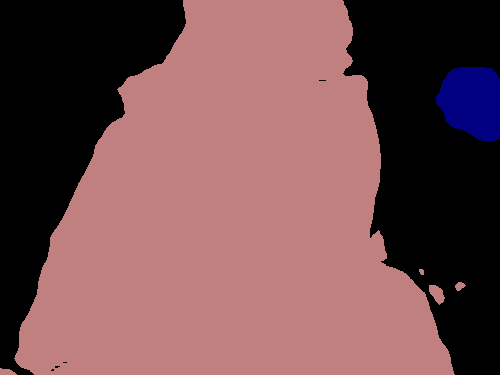}
\includegraphics[width=0.16\linewidth, height=3.5cm]{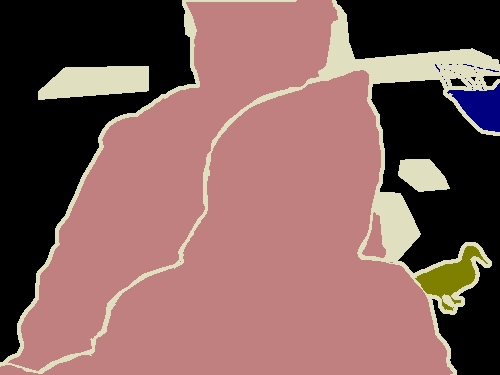}
\includegraphics[width=0.16\linewidth, height=3.5cm]{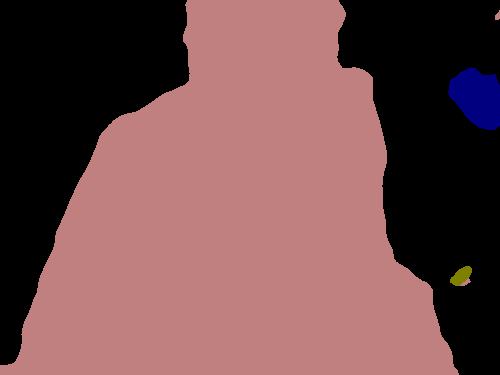}
\end{subfigure}

\begin{subfigure}{}
\includegraphics[width=0.16\linewidth, height=3.5cm]{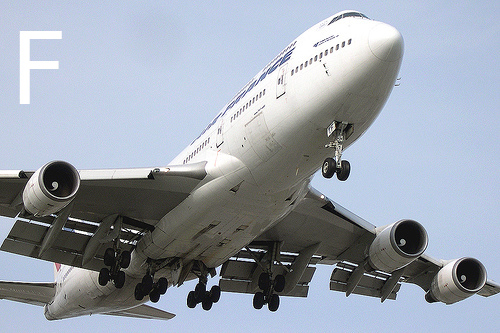} 
\includegraphics[width=0.16\linewidth, height=3.5cm]{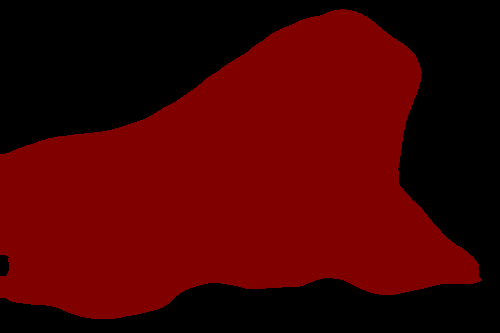} 
\includegraphics[width=0.16\linewidth, height=3.5cm]{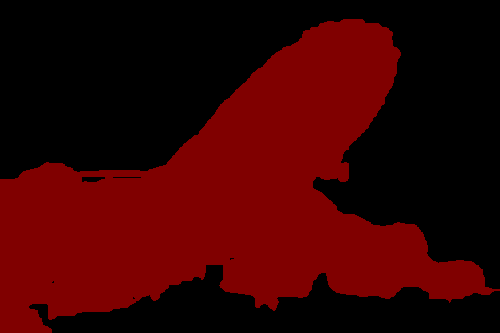} 
\includegraphics[width=0.16\linewidth, height=3.5cm]{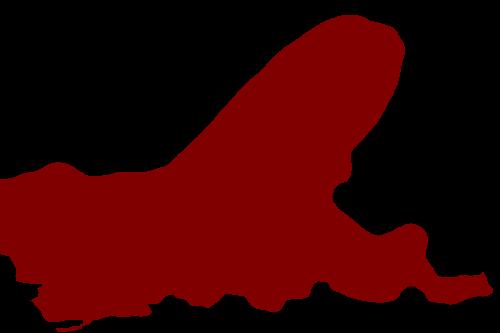}
\includegraphics[width=0.16\linewidth, height=3.5cm]{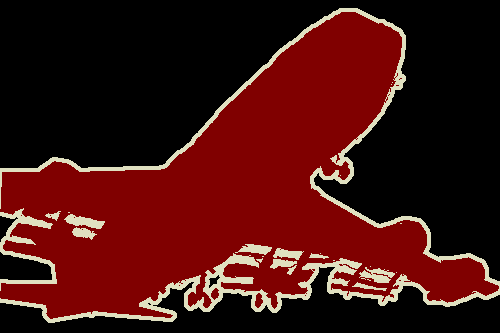}
\includegraphics[width=0.16\linewidth, height=3.5cm]{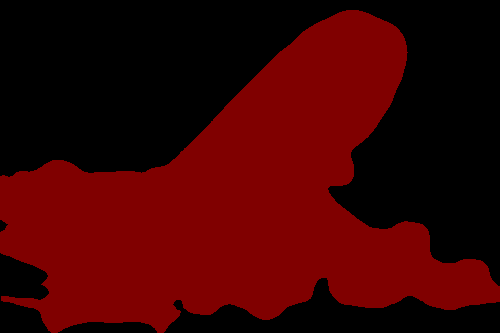}
\end{subfigure}
 
\caption{From left to right: original images, PCAM labels, refined PCAM labels, predictions from PCAM-supervised DeepLabv3+, ground truth segmentation, and predictions from fully supervised DeepLabv3+ for each image}
\label{fig:maingrid}
\end{figure*}

\begin{table}
\centering
\begin{tabular}{l|l|l} 
\hline
Method       & Sup.       & mIOU  \\ 
\hline
WhatsPoint \cite{bearman2016s}  & \textit{P} & 46.1  \\
\hline
MIL-FCN    \cite{pathak2014fully}      & \textit{I} & 25.7  \\ 
CCNN  \cite{pathak2015constrained} & \textit{I} & 35.3 \\
EM-Adapt    \cite{papandreou2015weakly} & \textit{I} & 38.2  \\
MIL+seg \cite{pinheiro2015image}     & \textit{I}  & 42.0 \\
DCSM    \cite{shimoda2016distinct}      & \textit{I} & 44.1  \\ 
BFBP    \cite{saleh2016built}      & \textit{I} & 46.6  \\ 
SEC    \cite{kolesnikov2016seed}      & \textit{I} & 50.7  \\ 
AF-SS    \cite{qi2016augmented}      & \textit{I} & 52.6  \\ 
Combining Cues    \cite{roy2017combining}      & \textit{I} & 52.8  \\ 
AE-PSL    \cite{wei2017object}      & \textit{I} & 55.0  \\ 
DSRG \cite{huang2018weakly} & \textit{I} & 61.4 \\
AffinityNet \cite{ahn2018learning}  & \textit{I} & 61.7  \\
IRNet   \cite{ahn2019weakly}     & \textit{I} & 63.5  \\
FickleNet \cite{lee2019ficklenet}   & \textit{I} & 64.9  \\
\hline
ScribbleSup \cite{lin2016scribblesup} & \textit{S} & 63.1 \\
NormCut \cite{tang2018normalized} & \textit{S} & 65.1 \\
\hline
BBox-seg    \cite{papandreou2015weakly} & \textit{B} & 60.6  \\ 
SDI    \cite{khoreva2017simple}      & \textit{B} & 65.7  \\ 
BCM    \cite{song2019box}      & \textit{B} & 66.8  \\ 
\hline
Ours - refined PCAM & \textit{P} & 68.5 \\
Ours - final          & \textit{P} & 70.5  \\ 
\hline
DeepLabv3+ & \textit{F} & 78.9  \\
\hline
\end{tabular}
\caption{mIOU comparison of recent or related weakly supervised semantic segmentation methods on the PASCAL VOC 2012 \textit{val} set. Sup. shows the level of supervision of each method where \textit{P} is point level, \textit{I} is image level, \textit{B} is bounding box level, \textit{S} is scribble level, and \textit{F} is fully supervised. Ours - final uses DeepLabv3+ trained on refined PCAMs for inference.}
\label{tab:miou}
\end{table}

\begin{table*}
\centering
\resizebox{\linewidth}{!}{%
\begin{tabular}{l|l|lllllllllllllllllllll|l} 
\hline
Method       & Sup.        & bkg           & aero          & bike          & bird          & boat          & bottle        & bus           & car           & cat           & chair         & cow           & table         & dog           & horse         & mkb           & person        & plant         & sheep         & sofa          & train         & tv            & mean            \\ 
\hline
EM-Adapt \cite{papandreou2015weakly}     & \textit{I}  & 67.2          & 29.2          & 17.6          & 28.6          & 22.2          & 29.6          & 47.0          & 44.0          & 44.2          & 14.6          & 35.1          & 24.9          & 41.0          & 34.8          & 41.6          & 32.1          & 24.8          & 37.4          & 24.0          & 38.1          & 31.6          & 33.8            \\
CCNN  \cite{pathak2015constrained}       & \textit{I}  & 68.5          & 25.5          & 18.0          & 25.4          & 20.2          & 36.3          & 46.8          & 47.1          & 48.0          & 15.8          & 37.9          & 21.0          & 44.5          & 34.5          & 46.2          & 40.7          & 30.4          & 36.3          & 22.2          & 38.8          & 36.9          & 35.3            \\
MIL+seg \cite{pinheiro2015image}     & \textit{I}  & 79.6          & 50.2          & 21.6          & 40.9          & 34.9          & 40.5          & 45.9          & 51.5          & 60.6          & 12.6          & 51.2          & 11.6          & 56.8          & 52.9          & 44.8          & 42.7          & 31.2          & 55.4          & 21.5          & 38.8          & 36.9          & 42.0            \\
SEC  \cite{kolesnikov2016seed}        & \textit{I}  & 82.4          & 62.9          & 26.4          & 61.6          & 27.6          & 38.1          & 66.6          & 62.7          & 75.2          & 22.1          & 53.5          & 28.3          & 65.8          & 57.8          & 62.3          & 52.5          & 32.5          & 62.6          & 32.1          & 45.4          & 45.3          & 50.7            \\
AE-PSL  \cite{wei2017object} & \textit{I}  & 83.4          & 71.1          & 30.5          & 72.9          & 41.6          & 55.9          & 63.1          & 60.2          & 74.0          & 18.0          & 66.5          & 32.4          & 71.7          & 56.3          & 64.8          & 52.4          & 37.4          & 69.1          & 31.4          & 58.9          & 43.9          & 55.0            \\
What's Point \cite{bearman2016s} & \textit{P}  & 80            & 49            & 23            & 39            & 41            & 46            & 60            & 61            & 56            & 18            & 38            & 41            & 54            & 42            & 55            & 57            & 32            & 51            & 26            & 55            & 45            & 46.0            \\
AffinityNet \cite{ahn2018learning}  & \textit{I}  & 88.2          & 68.2          & \textbf{30.6} & 81.1          & 49.6          & 61.0          & 77.8          & 66.1          & 75.1          & 29.0          & 66.0          & 40.2          & 80.4          & 62.0          & 70.4          & \textbf{73.7} & 42.5          & 70.7          & 42.6          & 68.1          & 51.6          & 61.7            \\
BCM      \cite{song2019box}    & \textit{B}  & 89.8          & \textbf{68.3} & 27.1          & 73.7          & 56.4          & 72.6          & 84.2          & 75.6          & 79.9          & \textbf{35.2} & 78.3          & 53.2          & 77.6          & 66.4          & 68.1          & 73.1          & \textbf{56.8} & 80.1          & 45.1          & \textbf{74.7} & 54.6          & 66.8            \\ 
\hline
Ours - final           & \textit{P}  & \textbf{89.9} & 66.1          & 30.1          & \textbf{85.2} & \textbf{62.5} & \textbf{75.8} & \textbf{87.1} & \textbf{80.4} & \textbf{87.1} & 34.0          & \textbf{85.1} & \textbf{60.0} & \textbf{84.4} & \textbf{82.4} & \textbf{77.4} & 68.4          & 56.1          & \textbf{84.0} & \textbf{46.1} & 72.6          & \textbf{64.2} & \textbf{70.5 }  \\
\hline
\end{tabular}
}
\caption{per class mIOU comparsion of recent or related methods on the PASCAL VOC 2012 val set}
\label{tab:iou}
\end{table*}

\subsubsection{DeepLabv3+}
We train DeepLabv3+ \cite{chen2018encoder} using the labels generated from refining PCAMs with IRNet. We use a ResNet backbone, a learning rate of 0.007, weight decay of 4e-5, and a Nesterov momentum optimizer with a momentum of 0.9. We use an output stride of 16 during training and evaluation, and we do not implement multi-scale or flipped inputs during training. This is the same training setting as DeepLabv3+ shown in Table \ref{tab:miou}; however, the performance shown there for "DeepLabv3+" is for the fully supervised setting trained on actual ground truth as opposed to "Ours - final" which is DeepLabv3+ trained on refined PCAMs.

\subsection{PCAM Performance}
Figure \ref{fig:camgrid} shows some localization performance of PCAMs compared to CAMs generated from the same network without point supervision. This figure shows that the point supervision generally leads to better localization. Further, PCAMs do a much better job of covering more of a given object's extent.

Table \ref{tab:CAM} shows the quantitative performance difference of each activation map. In addition to the image level CAM, we also experimented with training the network with the PCAM loss given every ground truth point, i.e. full supervision, rather than the much smaller one point per class per image. 
Point supervision strongly increases the localization performance of activation maps, though using all points in an image with our method seems to have relatively little influence on performance on unseen images.

\subsection{IRNet Label Refinement}

Figure \ref{fig:maingrid} shows several examples of our method's performance as well as the performance of the fully supervised DeepLabv3+. We can see evidence that the label propagation via IRNet's transition matrix usually helps refine boundaries more precisely, though we tend to see choppy edges as a result of the random walk propagation.

Table \ref{tab:refined} shows the performance difference of each activation map after having been refined by IRNet. Unsurprisingly, the already better PCAMs have significantly better localization performance after being refined than do CAMs. The performance increase between PCAMs and CAMs only becomes amplified after refinement. Before refinement, PCAMs outperform CAMs by 8.4\%, and after refinement, PCAMs surpasses CAMs by 11.1\%.

\subsection{Semantic Segmentation Results}

Figure \ref{fig:maingrid} shows the performance of DeepLabv3+ after being trained with refined PCAMs as well as its predictions after being trained with ground truth. The results are quite similar, with the PCAM-trained network generally making fewer predictions, which is helpful in the case of the image B where there is no false prediction in the bottom left corner. However, fewer predictions are not helpful in cases such as in the image E where the duck in the bottom right corner is not predicted at all. 

While the PCAM-trained CNN makes fewer predictions and seems to sometimes miss smaller objects, it adheres to boundaries better and performs slightly better quantitatively than the refined PCAMs themselves.

Table \ref{tab:miou} shows the performance of several recent methods on semantic segmentation of the VOC \textit{val} set. Our method achieves state of the art performance in point supervision. Further, it surpasses the method from Tang et al \cite{tang2018normalized}, which uses stronger scribble level supervision, and it even outperforms the very recent method by Song et al \cite{song2019box}, which uses stronger bounding box supervision for this task. Our method recovers an impressive 89\% of the performance of its fully supervised counterpart using only point supervision.

Table \ref{tab:iou} shows a number of recent methods, their levels of supervision, and their class-wise IOU performance. Unsurprisingly, the better supervised method from Song et al \cite{song2019box} performs better on the some of the challenging classes like chair; however, our method performs better on nearly every class and overall.

\section{Conclusion}
While numerous methods for using image-level supervision for weakly supervised semantic segmentation have been introduced recently, using point supervision has been largely unexplored. We propose a new method for training a class activation network to include point supervision. We demonstrate that this approach greatly enhances class activation maps, and we achieve state of the art performance for point supervised semantic segmentation that is even better than the state of the art methods using the stronger bounding box supervision. We achieve this performance using today's state of the art methods for label propagation over activation maps, but our method and PCAMs can be used in other pipelines as well.

{\small
\bibliographystyle{ieee_fullname}
\bibliography{egbib}

\begin{thebibliography}{10}\itemsep=-1pt

\bibitem{ahn2019weakly}
Jiwoon Ahn, Sunghyun Cho, and Suha Kwak.
\newblock Weakly supervised learning of instance segmentation with inter-pixel
  relations.
\newblock In {\em Proceedings of the IEEE Conference on Computer Vision and
  Pattern Recognition}, pages 2209--2218, 2019.

\bibitem{ahn2018learning}
Jiwoon Ahn and Suha Kwak.
\newblock Learning pixel-level semantic affinity with image-level supervision
  for weakly supervised semantic segmentation.
\newblock In {\em Proceedings of the IEEE Conference on Computer Vision and
  Pattern Recognition}, pages 4981--4990, 2018.

\bibitem{alexe2012measuring}
Bogdan Alexe, Thomas Deselaers, and Vittorio Ferrari.
\newblock Measuring the objectness of image windows.
\newblock {\em IEEE transactions on pattern analysis and machine intelligence},
  34(11):2189--2202, 2012.

\bibitem{bazzani2016self}
Loris Bazzani, Alessandra Bergamo, Dragomir Anguelov, and Lorenzo Torresani.
\newblock Self-taught object localization with deep networks.
\newblock In {\em 2016 IEEE winter conference on applications of computer
  vision (WACV)}, pages 1--9. IEEE, 2016.

\bibitem{bearman2016s}
Amy Bearman, Olga Russakovsky, Vittorio Ferrari, and Li Fei-Fei.
\newblock What’s the point: Semantic segmentation with point supervision.
\newblock In {\em European conference on computer vision}, pages 549--565.
  Springer, 2016.

\bibitem{briq2018convolutional}
Rania Briq, Michael Moeller, and J{\"u}rgen Gall.
\newblock Convolutional simplex projection network for weakly supervised
  semantic segmentation.
\newblock In {\em BMVC}, page 263, 2018.

\bibitem{chattopadhay2018grad}
Aditya Chattopadhay, Anirban Sarkar, Prantik Howlader, and Vineeth~N
  Balasubramanian.
\newblock Grad-cam++: Generalized gradient-based visual explanations for deep
  convolutional networks.
\newblock In {\em 2018 IEEE Winter Conference on Applications of Computer
  Vision (WACV)}, pages 839--847. IEEE, 2018.

\bibitem{chen2018encoder}
Liang-Chieh Chen, Yukun Zhu, George Papandreou, Florian Schroff, and Hartwig
  Adam.
\newblock Encoder-decoder with atrous separable convolution for semantic image
  segmentation.
\newblock In {\em Proceedings of the European conference on computer vision
  (ECCV)}, pages 801--818, 2018.

\bibitem{cholakkal2016backtracking}
Hisham Cholakkal, Jubin Johnson, and Deepu Rajan.
\newblock Backtracking scspm image classifier for weakly supervised top-down
  saliency.
\newblock In {\em Proceedings of the IEEE Conference on Computer Vision and
  Pattern Recognition}, pages 5278--5287, 2016.

\bibitem{cinbis2016weakly}
Ramazan~Gokberk Cinbis, Jakob Verbeek, and Cordelia Schmid.
\newblock Weakly supervised object localization with multi-fold multiple
  instance learning.
\newblock {\em IEEE transactions on pattern analysis and machine intelligence},
  39(1):189--203, 2016.

\bibitem{dai2015boxsup}
Jifeng Dai, Kaiming He, and Jian Sun.
\newblock Boxsup: Exploiting bounding boxes to supervise convolutional networks
  for semantic segmentation.
\newblock In {\em Proceedings of the IEEE International Conference on Computer
  Vision}, pages 1635--1643, 2015.

\bibitem{durand2017wildcat}
Thibaut Durand, Taylor Mordan, Nicolas Thome, and Matthieu Cord.
\newblock Wildcat: Weakly supervised learning of deep convnets for image
  classification, pointwise localization and segmentation.
\newblock In {\em Proceedings of the IEEE conference on computer vision and
  pattern recognition}, pages 642--651, 2017.

\bibitem{everingham2010pascal}
Mark Everingham, Luc Van~Gool, Christopher~KI Williams, John Winn, and Andrew
  Zisserman.
\newblock The pascal visual object classes (voc) challenge.
\newblock {\em International journal of computer vision}, 88(2):303--338, 2010.

\bibitem{ge2018multi}
Weifeng Ge, Sibei Yang, and Yizhou Yu.
\newblock Multi-evidence filtering and fusion for multi-label classification,
  object detection and semantic segmentation based on weakly supervised
  learning.
\newblock In {\em Proceedings of the IEEE Conference on Computer Vision and
  Pattern Recognition}, pages 1277--1286, 2018.

\bibitem{hariharan2011semantic}
Bharath Hariharan, Pablo Arbel{\'a}ez, Lubomir Bourdev, Subhransu Maji, and
  Jitendra Malik.
\newblock Semantic contours from inverse detectors.
\newblock In {\em 2011 International Conference on Computer Vision}, pages
  991--998. IEEE, 2011.

\bibitem{he2016deep}
Kaiming He, Xiangyu Zhang, Shaoqing Ren, and Jian Sun.
\newblock Deep residual learning for image recognition.
\newblock In {\em Proceedings of the IEEE conference on computer vision and
  pattern recognition}, pages 770--778, 2016.

\bibitem{hong2017weakly}
Seunghoon Hong, Donghun Yeo, Suha Kwak, Honglak Lee, and Bohyung Han.
\newblock Weakly supervised semantic segmentation using web-crawled videos.
\newblock In {\em Proceedings of the IEEE Conference on Computer Vision and
  Pattern Recognition}, pages 7322--7330, 2017.

\bibitem{hou2007saliency}
Xiaodi Hou and Liqing Zhang.
\newblock Saliency detection: A spectral residual approach.
\newblock In {\em 2007 IEEE Conference on Computer Vision and Pattern
  Recognition}, pages 1--8. Ieee, 2007.

\bibitem{hu2018learning}
Ronghang Hu, Piotr Doll{\'a}r, Kaiming He, Trevor Darrell, and Ross Girshick.
\newblock Learning to segment every thing.
\newblock In {\em Proceedings of the IEEE Conference on Computer Vision and
  Pattern Recognition}, pages 4233--4241, 2018.

\bibitem{huang2018weakly}
Zilong Huang, Xinggang Wang, Jiasi Wang, Wenyu Liu, and Jingdong Wang.
\newblock Weakly-supervised semantic segmentation network with deep seeded
  region growing.
\newblock In {\em Proceedings of the IEEE Conference on Computer Vision and
  Pattern Recognition}, pages 7014--7023, 2018.

\bibitem{itti1998model}
Laurent Itti, Christof Koch, and Ernst Niebur.
\newblock A model of saliency-based visual attention for rapid scene analysis.
\newblock {\em IEEE Transactions on Pattern Analysis \& Machine Intelligence},
  (11):1254--1259, 1998.

\bibitem{khoreva2017simple}
Anna Khoreva, Rodrigo Benenson, Jan Hosang, Matthias Hein, and Bernt Schiele.
\newblock Simple does it: Weakly supervised instance and semantic segmentation.
\newblock In {\em Proceedings of the IEEE conference on computer vision and
  pattern recognition}, pages 876--885, 2017.

\bibitem{kolesnikov2016seed}
Alexander Kolesnikov and Christoph~H Lampert.
\newblock Seed, expand and constrain: Three principles for weakly-supervised
  image segmentation.
\newblock In {\em European Conference on Computer Vision}, pages 695--711.
  Springer, 2016.

\bibitem{kwak2017weakly}
Suha Kwak, Seunghoon Hong, and Bohyung Han.
\newblock Weakly supervised semantic segmentation using superpixel pooling
  network.
\newblock In {\em Thirty-First AAAI Conference on Artificial Intelligence},
  2017.

\bibitem{lee2019ficklenet}
Jungbeom Lee, Eunji Kim, Sungmin Lee, Jangho Lee, and Sungroh Yoon.
\newblock Ficklenet: Weakly and semi-supervised semantic image segmentation
  using stochastic inference.
\newblock In {\em Proceedings of the IEEE Conference on Computer Vision and
  Pattern Recognition}, pages 5267--5276, 2019.

\bibitem{liang2016learning}
Xiaodan Liang, Yunchao Wei, Liang Lin, Yunpeng Chen, Xiaohui Shen, Jianchao
  Yang, and Shuicheng Yan.
\newblock Learning to segment human by watching youtube.
\newblock {\em IEEE transactions on pattern analysis and machine intelligence},
  39(7):1462--1468, 2016.

\bibitem{lin2016scribblesup}
Di Lin, Jifeng Dai, Jiaya Jia, Kaiming He, and Jian Sun.
\newblock Scribblesup: Scribble-supervised convolutional networks for semantic
  segmentation.
\newblock In {\em Proceedings of the IEEE Conference on Computer Vision and
  Pattern Recognition}, pages 3159--3167, 2016.

\bibitem{lu2016learning}
Zhiwu Lu, Zhenyong Fu, Tao Xiang, Peng Han, Liwei Wang, and Xin Gao.
\newblock Learning from weak and noisy labels for semantic segmentation.
\newblock {\em IEEE transactions on pattern analysis and machine intelligence},
  39(3):486--500, 2016.

\bibitem{maninis2018deep}
Kevis-Kokitsi Maninis, Sergi Caelles, Jordi Pont-Tuset, and Luc Van~Gool.
\newblock Deep extreme cut: From extreme points to object segmentation.
\newblock In {\em Proceedings of the IEEE Conference on Computer Vision and
  Pattern Recognition}, pages 616--625, 2018.

\bibitem{oquab2014learning}
Maxime Oquab, Leon Bottou, Ivan Laptev, and Josef Sivic.
\newblock Learning and transferring mid-level image representations using
  convolutional neural networks.
\newblock In {\em Proceedings of the IEEE conference on computer vision and
  pattern recognition}, pages 1717--1724, 2014.

\bibitem{oquab2015object}
Maxime Oquab, L{\'e}on Bottou, Ivan Laptev, and Josef Sivic.
\newblock Is object localization for free?-weakly-supervised learning with
  convolutional neural networks.
\newblock In {\em Proceedings of the IEEE Conference on Computer Vision and
  Pattern Recognition}, pages 685--694, 2015.

\bibitem{papandreou2015weakly}
George Papandreou, Liang-Chieh Chen, Kevin~P Murphy, and Alan~L Yuille.
\newblock Weakly-and semi-supervised learning of a deep convolutional network
  for semantic image segmentation.
\newblock In {\em Proceedings of the IEEE international conference on computer
  vision}, pages 1742--1750, 2015.

\bibitem{pathak2015constrained}
Deepak Pathak, Philipp Krahenbuhl, and Trevor Darrell.
\newblock Constrained convolutional neural networks for weakly supervised
  segmentation.
\newblock In {\em Proceedings of the IEEE international conference on computer
  vision}, pages 1796--1804, 2015.

\bibitem{pathak2014fully}
Deepak Pathak, Evan Shelhamer, Jonathan Long, and Trevor Darrell.
\newblock Fully convolutional multi-class multiple instance learning.
\newblock {\em arXiv preprint arXiv:1412.7144}, 2014.

\bibitem{pinheiro2015image}
Pedro~O Pinheiro and Ronan Collobert.
\newblock From image-level to pixel-level labeling with convolutional networks.
\newblock In {\em Proceedings of the IEEE Conference on Computer Vision and
  Pattern Recognition}, pages 1713--1721, 2015.

\bibitem{qi2016augmented}
Xiaojuan Qi, Zhengzhe Liu, Jianping Shi, Hengshuang Zhao, and Jiaya Jia.
\newblock Augmented feedback in semantic segmentation under image level
  supervision.
\newblock In {\em European Conference on Computer Vision}, pages 90--105.
  Springer, 2016.

\bibitem{rajchl2016deepcut}
Martin Rajchl, Matthew~CH Lee, Ozan Oktay, Konstantinos Kamnitsas, Jonathan
  Passerat-Palmbach, Wenjia Bai, Mellisa Damodaram, Mary~A Rutherford, Joseph~V
  Hajnal, Bernhard Kainz, et~al.
\newblock Deepcut: Object segmentation from bounding box annotations using
  convolutional neural networks.
\newblock {\em IEEE transactions on medical imaging}, 36(2):674--683, 2016.

\bibitem{roy2017combining}
Anirban Roy and Sinisa Todorovic.
\newblock Combining bottom-up, top-down, and smoothness cues for weakly
  supervised image segmentation.
\newblock In {\em Proceedings of the IEEE Conference on Computer Vision and
  Pattern Recognition}, pages 3529--3538, 2017.

\bibitem{saleh2016built}
Fatemehsadat Saleh, Mohammad~Sadegh Aliakbarian, Mathieu Salzmann, Lars
  Petersson, Stephen Gould, and Jose~M Alvarez.
\newblock Built-in foreground/background prior for weakly-supervised semantic
  segmentation.
\newblock In {\em European Conference on Computer Vision}, pages 413--432.
  Springer, 2016.

\bibitem{selvaraju2017grad}
Ramprasaath~R Selvaraju, Michael Cogswell, Abhishek Das, Ramakrishna Vedantam,
  Devi Parikh, and Dhruv Batra.
\newblock Grad-cam: Visual explanations from deep networks via gradient-based
  localization.
\newblock In {\em Proceedings of the IEEE International Conference on Computer
  Vision}, pages 618--626, 2017.

\bibitem{shimoda2016distinct}
Wataru Shimoda and Keiji Yanai.
\newblock Distinct class-specific saliency maps for weakly supervised semantic
  segmentation.
\newblock In {\em European Conference on Computer Vision}, pages 218--234.
  Springer, 2016.

\bibitem{song2019box}
Chunfeng Song, Yan Huang, Wanli Ouyang, and Liang Wang.
\newblock Box-driven class-wise region masking and filling rate guided loss for
  weakly supervised semantic segmentation.
\newblock In {\em Proceedings of the IEEE Conference on Computer Vision and
  Pattern Recognition}, pages 3136--3145, 2019.

\bibitem{tang2018normalized}
Meng Tang, Abdelaziz Djelouah, Federico Perazzi, Yuri Boykov, and Christopher
  Schroers.
\newblock Normalized cut loss for weakly-supervised cnn segmentation.
\newblock In {\em Proceedings of the IEEE Conference on Computer Vision and
  Pattern Recognition}, pages 1818--1827, 2018.

\bibitem{vernaza2017learning}
Paul Vernaza and Manmohan Chandraker.
\newblock Learning random-walk label propagation for weakly-supervised semantic
  segmentation.
\newblock In {\em Proceedings of the IEEE conference on computer vision and
  pattern recognition}, pages 7158--7166, 2017.

\bibitem{wang2018weakly}
Xiang Wang, Shaodi You, Xi Li, and Huimin Ma.
\newblock Weakly-supervised semantic segmentation by iteratively mining common
  object features.
\newblock In {\em Proceedings of the IEEE conference on computer vision and
  pattern recognition}, pages 1354--1362, 2018.

\bibitem{wei2017object}
Yunchao Wei, Jiashi Feng, Xiaodan Liang, Ming-Ming Cheng, Yao Zhao, and
  Shuicheng Yan.
\newblock Object region mining with adversarial erasing: A simple
  classification to semantic segmentation approach.
\newblock In {\em Proceedings of the IEEE conference on computer vision and
  pattern recognition}, pages 1568--1576, 2017.

\bibitem{wei2018revisiting}
Yunchao Wei, Huaxin Xiao, Honghui Shi, Zequn Jie, Jiashi Feng, and Thomas~S
  Huang.
\newblock Revisiting dilated convolution: A simple approach for weakly-and
  semi-supervised semantic segmentation.
\newblock In {\em Proceedings of the IEEE Conference on Computer Vision and
  Pattern Recognition}, pages 7268--7277, 2018.

\bibitem{xu2015learning}
Jia Xu, Alexander~G Schwing, and Raquel Urtasun.
\newblock Learning to segment under various forms of weak supervision.
\newblock In {\em Proceedings of the IEEE conference on computer vision and
  pattern recognition}, pages 3781--3790, 2015.

\bibitem{zhang2018adversarial}
Xiaolin Zhang, Yunchao Wei, Jiashi Feng, Yi Yang, and Thomas~S Huang.
\newblock Adversarial complementary learning for weakly supervised object
  localization.
\newblock In {\em Proceedings of the IEEE Conference on Computer Vision and
  Pattern Recognition}, pages 1325--1334, 2018.

\bibitem{zhao2018pseudo}
Xiangyun Zhao, Shuang Liang, and Yichen Wei.
\newblock Pseudo mask augmented object detection.
\newblock In {\em Proceedings of the IEEE conference on computer vision and
  pattern recognition}, pages 4061--4070, 2018.

\bibitem{zhou2016learning}
Bolei Zhou, Aditya Khosla, Agata Lapedriza, Aude Oliva, and Antonio Torralba.
\newblock Learning deep features for discriminative localization.
\newblock In {\em Proceedings of the IEEE conference on computer vision and
  pattern recognition}, pages 2921--2929, 2016.

\end{thebibliography}
}

\end{document}